\documentclass{article}

\usepackage[a4paper, total={6in, 9in}]{geometry}
\usepackage{hyperref}
\usepackage[T1]{fontenc}
\usepackage{lmodern}

\usepackage{times}
\usepackage{graphicx}
\usepackage{latexsym}
\usepackage{subfigure}
\usepackage{booktabs}
\usepackage{amsmath,amssymb}
\usepackage{url}
\usepackage{authblk}
\usepackage{multirow}
\usepackage{array}
\newcommand\MyBox[2]{
  \fbox{\lower0.75cm
    \vbox to 1.7cm{\vfil
      \hbox to 1.7cm{\hfil\parbox{1.4cm}{#1\\#2}\hfil}
      \vfil}%
  }%
}

\newcommand{\vct}[1]{\boldsymbol{#1}} % vector
\newcommand{\mat}[1]{\boldsymbol{#1}} % matrix

\newcommand{\field}[1]{\mathbb{#1}}
\newcommand{\R}{\field{R}} % real domain
\DeclareMathOperator{\argmax}{arg\,max}

\usepackage{crimson}
\begin{document}

\title{\LARGE Precipitation Nowcasting with Star-Bridge Networks}

\author[1]{Yuan Cao}
\author[1]{Qiuying Li}
\author[1]{Hongming Shan}
\author[1]{Zhizhong Huang}
\author[2]{Lei Chen}
\author[2]{Leiming Ma}
\author[1]{Junping Zhang}
\affil[1]{Shanghai Key Laboratory of Intelligent Information Processing,  School of Computer Science, Fudan University, Shanghai 200433,    China}
\affil[2]{Shanghai Observatory, Shanghai, China}
\maketitle

\begin{abstract}
Precipitation nowcasting, which aims to precisely predict the short-term rainfall intensity of a local region, is gaining increasing attention in the artificial intelligence community. Existing deep learning-based algorithms use a single network to process various rainfall intensities together, compromising the predictive accuracy. Therefore, this paper proposes a novel recurrent neural network (RNN) based star-bridge network ({\bf StarBriNet}) for precipitation nowcasting. The novelty of this work lies in the following three aspects. First, the proposed network comprises multiple sub-networks to deal with different rainfall intensities and duration separately, which can significantly improve the model performance. Second, we propose a star-shaped information bridge to enhance the information flow across RNN layers. Third, we introduce a multi-sigmoid loss function to take the precipitation nowcasting criterion into account. Experimental results demonstrate superior performance for precipitation nowcasting over existing algorithms, including the state-of-the-art one, on a natural radar echo dataset.
\end{abstract}

\section{Introduction}

Nowcasting precipitation plays a crucial role in agriculture, flood alerting, daily life of citizens,  transportation, and so on, which is gaining increasing attention in the artificial intelligence community. Predicting the short-term rainfall intensity in a region is challenging as it relies on a number of meteorological factors such as temperature, humidity, wind, pressure in the region enclosing the clouds and the ground surface. For example, the involved convective precipitation occurs in the form of showers of high intensity and short duration, which is one of the most challenging precipitation forms. Traditional numerical weather prediction (NWP) models suffer from inferior prediction performance in terms of accuracy and resolution.

Given the huge amount of video-like radar echo data provided by the operational weather radar networks, the literature proposed several deep learning-based methods in the form of video prediction for precipitation nowcasting. With the consecutive 6-minute-interval radar echo frames as the input and output data, Shi~\emph{et. al} first dealt with the issue of precipitation nowcasting based on the framework of recurrent neural network(RNN)~\cite{convlstm}. More specifically, they  proposed encoder-decoder based convolutional long short-term memory (ConvLSTM), which is broadly used in various video tasks. Then Shi~\emph{et. al}~\cite{shi2017deep} utilized optical flow warping to refine the convolutional region of ConvLSTM. Remarkably, predictive RNN (PredRNN)~\cite{wang2017predrnn} and its variant PredRNN++~\cite{predrnnpp} achieved the state-of-the-art performance by adding a zigzag memory connection across different LSTM layers.

\begin{figure}[t]
\includegraphics[width=1\linewidth]{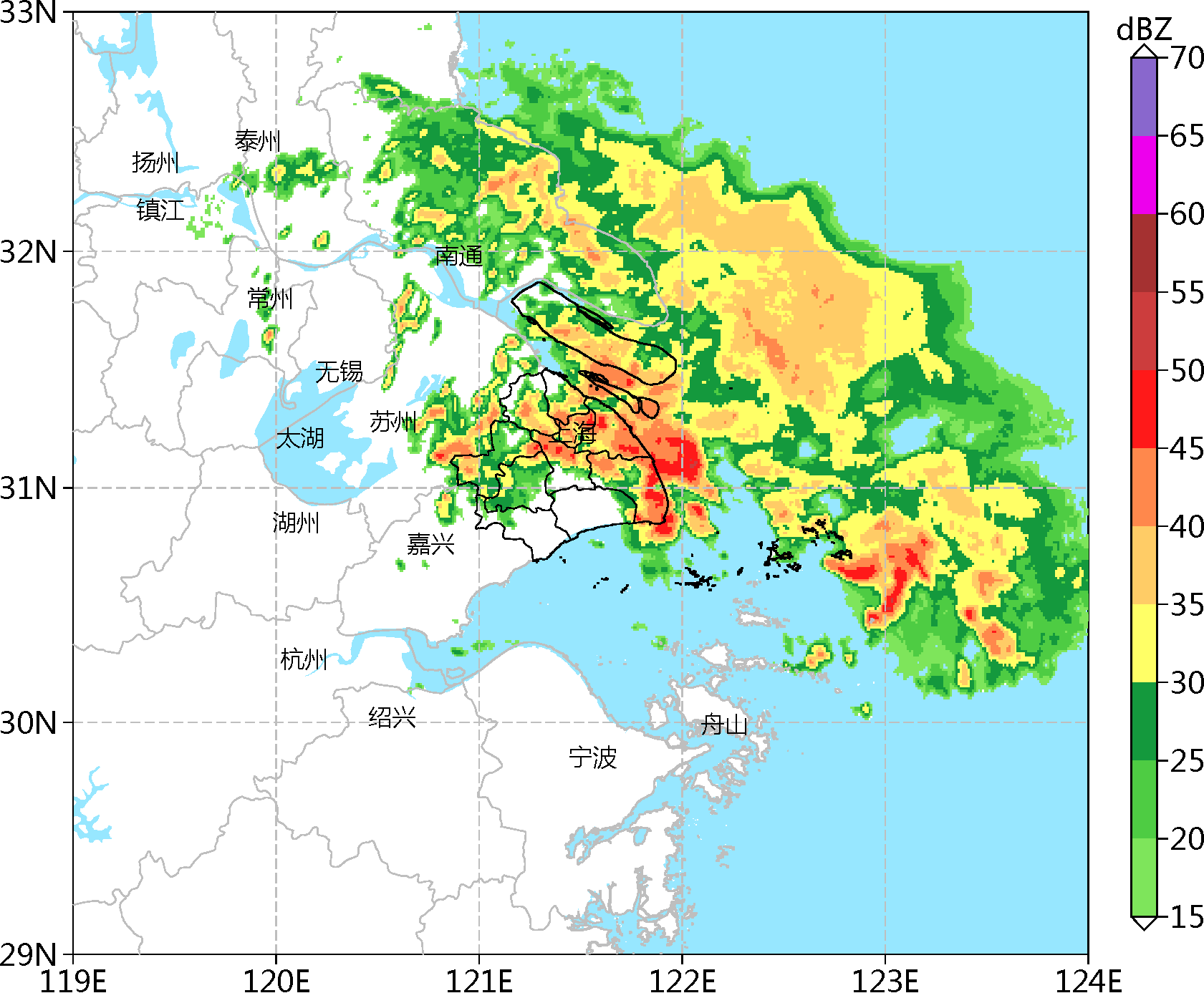}
\vspace{-15pt}
\caption{ The radar echo frame overlaid on the map of east China showing the current precipitation in terms of dBZ.} \label{fig:weather}
\end{figure}

Although video prediction methods can be directly applied to the precipitation nowcasting task, we highlight that the precipitation nowcasting has following significant differences from video prediction. First, precipitation is highly influenced by temperature, atmosphere, wind, humidity and others apart from radar echo images data, which means that precipitation nowcasting is more complicated than video prediction; see Fig.~\ref{fig:weather}. Second, precipitation has various forms in terms of different intensities and duration, which is more challenging. Third, in addition to nowcasting precipitation, weather criterion such as critical success index (CSI) indicates that the predicted rainfall intensity above some predefined threshold is more important than those below it. Currently existing methods, however, directly adopted the video prediction methods for precipitation nowcasting, which compromises the prediction performance.

To address those problems, this paper proposes a novel recurrent neural network (RNN) based star-bridge network ({\bf StarBriNet}) for precipitation nowcasting. More specifically, our proposed network 1) includes multi-column structure to process different rainfall intensities and duration separately; 2) contains a star-shaped information bridge to enhance the information flow across RNN layers; and 3) uses a novel multi-sigmoid loss function to optimize the parameters by taking weather criterion into account. We evaluate our StarBriNet on a Radar Echo dataset of east China in comparison with other baseline algorithms, including the state-of-the-art one.

\begin{figure*}[t]
    \centering
    \includegraphics[width=0.8\linewidth]{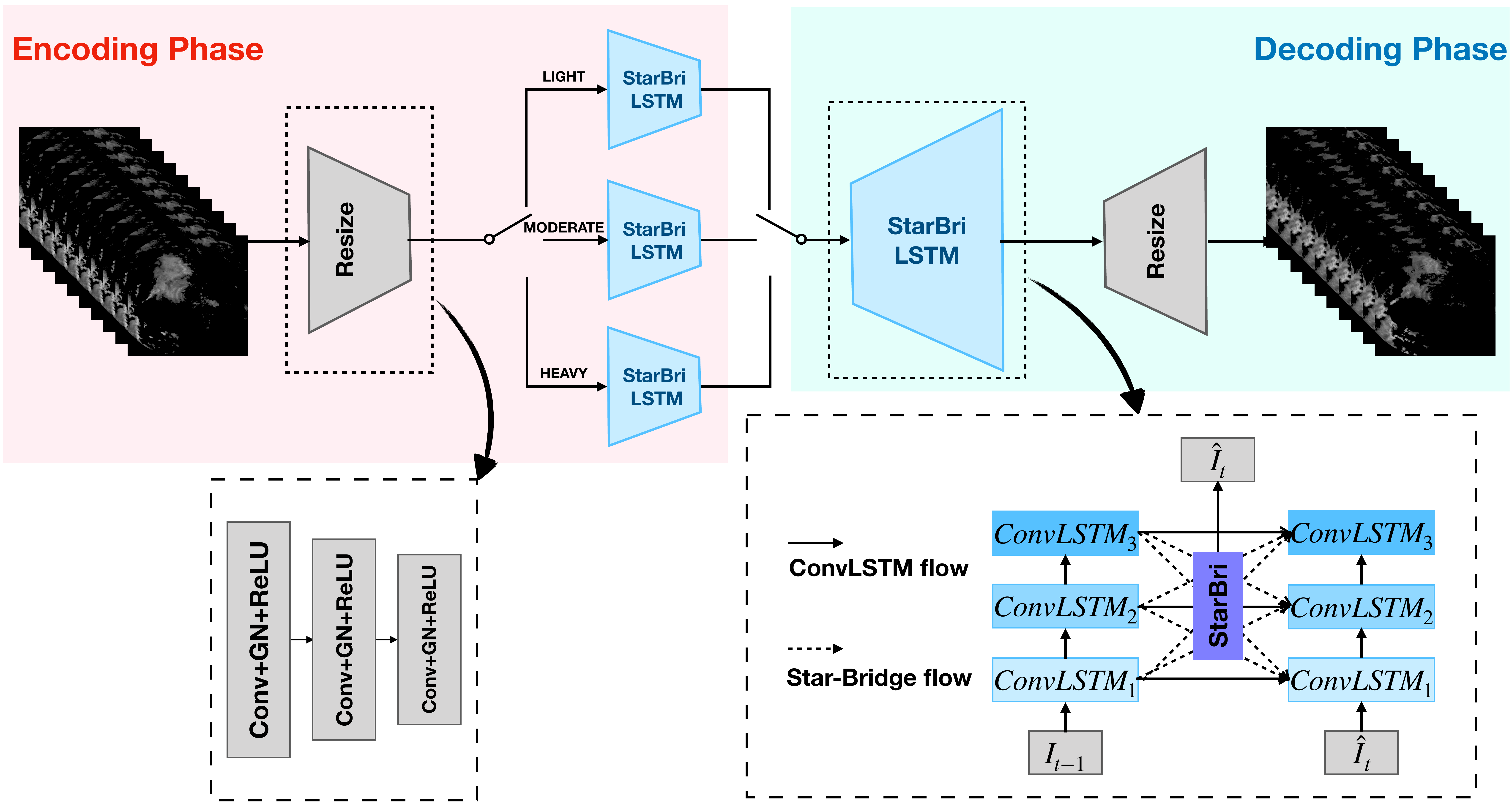}
    \caption{Overall architecture of our Star-Bridge Network} \label{fig:network}
\end{figure*}
The contributions of this paper are summarized as follows.
\begin{enumerate}
\item We propose to use multi-column RNNs to deal with different intensities and duration of the rainfall separately.
\item We propose a star-shaped information bridge to enhance the information flow across RNN layers.
\item We introduce a new multi-sigmoid loss to take the precipitation nowcasting criterion into account.
\end{enumerate}

\section{Preliminaries} \label{Related Works}

Compared to video prediction tasks, the precipitation nowcasting task of radar echo data is more difficult and complicated because the `rain clouds' on radar echo images vary in the moving speed, density, and contour.

\paragraph{Problem definition}
Let $\mat{S}_{1:T} = [\mat{I}_1, \mat{I}_2, \ldots, \mat{I}_T]$ be a radar echo sequence of length $T$, with each frame $\mat{I}_t \in \R^{H \times W}$ of size $H \times W$, $\forall t \in \{ 1,\ldots, T \}$. The goal of precipitation nowcasting is to predict the future radar echo sequence of length $L>0$ subsequent to current radar echo sequence $1:T$, which can be formulated as follows:
\begin{equation}
\widehat{\mat{S}}_{T+1:T+L} = \argmax_{\mat{S}_{T+1:T+L}} p(\mat{S}_{T+1:T+L} | \mat{S}_{1:T})
\end{equation}
where $\widehat{\mat{S}}_{T+1:T+L}$ represents the  predicted radar echo sequence of length $L$.

\paragraph{Convolutional LSTMs}
Significant progress has been made with ConvLSTMs towards video-based tasks like action detection~\cite{Song_2019_CVPR}, video object detection ~\cite{song2018pyramid,Li_2018_CVPR}, and video prediction~\cite{predrnnpp, wang2018eidetic} since 2015. Apart from CNN-based methods, researchers also dig into RNN-based methods for video prediction. Convolutional LSTM~\cite{convlstm} is an effective one among various RNN methods. Following the long short-term memory model, ConvLSTM utilizes its recurrent neural network architecture to memorize temporal information in a video sequence and extracts the spatial feature maps by using convolutional operation. We utilize a simple version of ConvLSTM, the key equations of ConvLSTM are summarized as follows:
\begin{align}
% \begin{equation}
\begin{split}
    \vct{f}_{t}, \vct{i}_{t}, \vct{o}_t &=\sigma(\mat{W}_{g}\circledast[\vct{x}_{t},\vct{h}_{t-1}]+\vct{b}_{g})\\
    \vct{c}_{t}&=\vct{f}_{t}\odot \vct{c}_{t-1}+\vct{i}_{t}\odot \mathrm{tanh}(\mat{W}_{c}\circledast[\vct{x}_t,\vct{h}_{t-1}]+\vct{b}_{c})\\
    \vct{h}_{t}&=\vct{o}_{t} \odot \mathrm{tanh}(\vct{c}_{t}),
\end{split}
% \end{equation}
\end{align}
where $\circledast$ denotes the convolution operator, $\odot$ denotes the element-wise product,  $\vct{b}$ is bias term of convolution kernel, $\vct{c}_{t-1}$ is the cell memory of last time-step, $\vct{h}_{t-1}$ is the output of last time-step, and $\sigma$ stands for the sigmoid function.

\section{Methods} \label{method}

In this section, we present our proposed StarBriNet in detail.
Subsections \ref{3.1}, \ref{3.2}, and \ref{msl} introduce the
Multi-Column RNNs, the star-shaped information bridge to enhance the
information flow across RNN layers,  and a multi-sigmoid loss
function to take the precipitation nowcasting criterion into
account, respectively. We use an encoder-decoder structure for
precipitation nowcasting shown in Fig.~\ref{fig:network}. Similar
to~\cite{srivastava2015unsupervised}, the initial cell memory and
cell output of decoding network are directly copied from the last
state of the encoding network. Input sequence is fed into a Resize
network consisting of 3 convolutional layers to reduce the size of
feature maps. Based on the computed mean and first-order derivative
(changing rate) of rain fall intensity, the output of the Resize
network passes through one of the three StarBri LSTM encoders that
correspond to light, moderate, and heavy rainfall intensities. We
call these three RNN encoders as multi-column RNNs. Then, the cell
states of StarBriLSTM encoder are transfered to a StarBri LSTM
decoder as the initial states. Another Resize network consisting of
three deconvolution layers up-samples the predicted frames to its
original size. Instead of widely-used loss functions $L_1$ and
$L_2$, we introduce a multi-sigmoid loss, which takes  precipitation
nowcasting criterion into account.

\begin{figure}[t]
\centering
\includegraphics[width=1\linewidth]{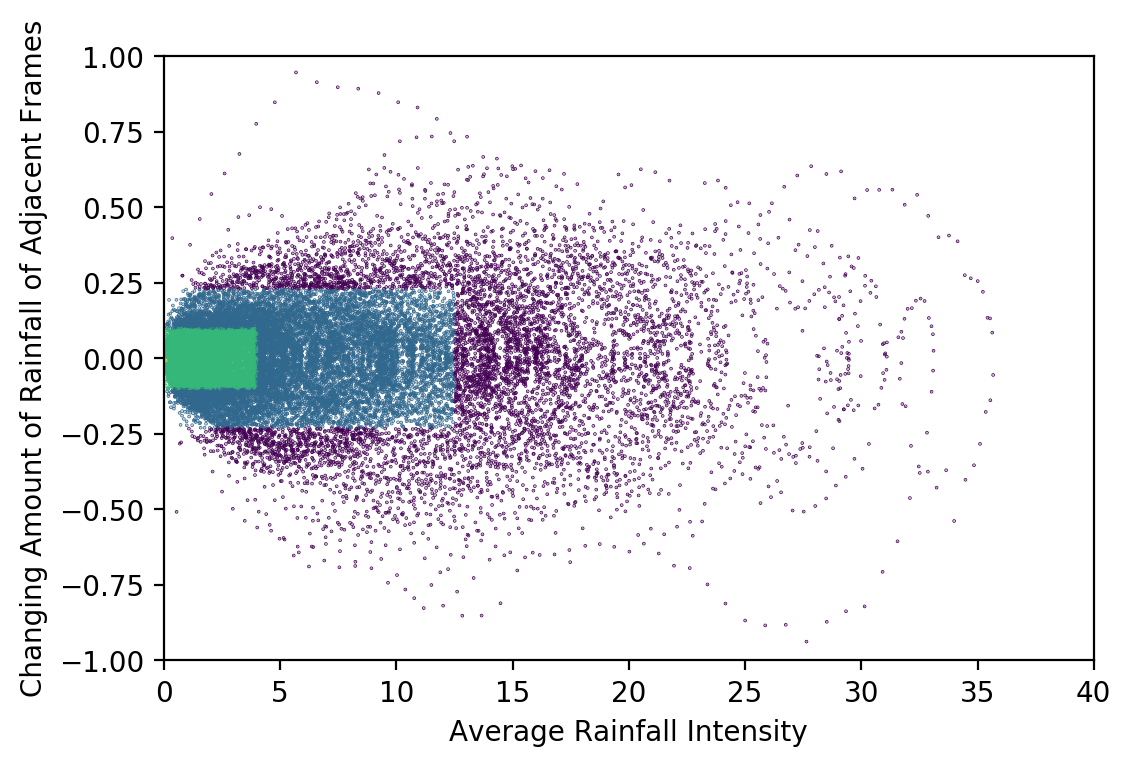}
\caption{The Distribution of samples in Radar Echo Dataset. 59.2\% samples fall in the little green box, 25.4\% for the blue one, and the other 15.4\% samples are purple.}
\label{fig:Distribution}
\end{figure}

\subsection{Multi-Column RNNs}\label{3.1}

We visualize the rainfall intensity and its changing rate of each sample sequence in Fig.~ \ref{fig:Distribution}. The diversity of data distribution is the key problem of precipitation nowcasting task. Therefore, we employ a divide-and-conquer strategy to address the diverse distribution. To this end, we propose a multi-column RNN structure in our encoder-decoder networks. Each column, which is an RNN encoder in this case, is explicitly designed to process one particular range of rainfall intensity.

As shown in Fig.~\ref{fig:network}, our proposed architecture consists of three StarBri-LSTM encoders for light, moderate, and heavy rainfall intensities, respectively. We separate the whole precipitation dataset into three classes with respect to the rainfall intensity and its first order derivative along time axis of each sequence sample. For such a data partition, rainfall intensity and changing rate in each class are less divergent. Each encoder consumes one particular intensity level in both training and testing phase to learn a specified circumstances, for example the light rain encoder only processes samples located in the small green box in Fig.~\ref{fig:Distribution}. Encoded features are transferred to StarBriLSTM decoder as the initial cell state. Furthermore, we also make explorations on 1-encoder 3-decoder and 3-encoder 3-decoder structures. The performances of these variants are slightly lower than the one in Fig.~\ref{fig:network}. Parameter numbers will be 3 times larger but the floating-point operations are almost the same as the single-column one.

\subsection{Star-Shaped Information Bridge} \label{3.2}

\begin{figure}[t]
\begin{minipage}[t]{1\linewidth}
\centering
\includegraphics[width=2.2in]{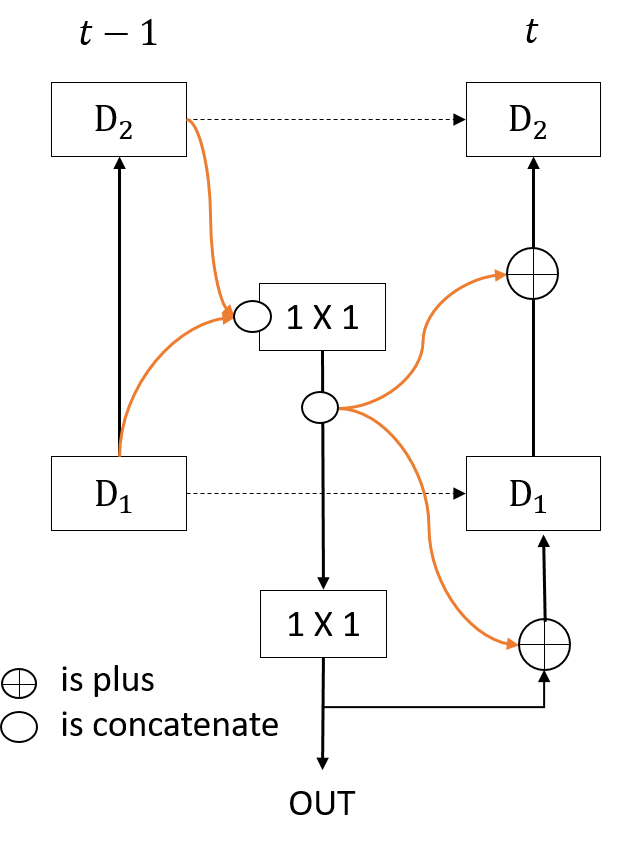}
\caption{The star-shaped information bridge in decoder network.}
\label{fig:star}
\end{minipage}
\end{figure}

We first utilize the Convolutional LSTM (ConvLSTM) network as our basic building block.
Traditional multi-layer LSTMs usually take the last output $\hat{I}_{t-1}$ as the input of the first layer at next time-step. A disadvantage of this strategy is that it may bring accumulative errors to predictions in testing stage, because $\hat{I}_{t-1}$ is different from the ground truth input $I_{t-1}$. But this connection passes more information across time steps which benefits the back-propagation, and this is the main reason why researcher still use it. We further expand feature transfer by introducing a novel Star-Shape information bridge to add more information from the last time-step to make the feature flow in multi-layer ConvLSTM more robust. More specifically, we concatenate output of all ConvLSTM layers and pass it to a $1\times1$ convolution layer and split the output of convolution layer to all ConvLSTM layers of next time-step by a residual connection to their inputs. Fig.~\ref{fig:star} demonstrate the star-shape information bridge binded to a 2 layer decoder.
It is hard to train the star-shape structure because of gradient exploding problem. Therefore, we attach the group normalization after each convolution layer and greatly relieve this hard-training problem. We also add group normalization layer to every convolution layer in ConvLSTM which directly improve our prediction performance by a large margin.

\subsection{Multi-Sigmoid Loss} \label{msl}

\begin{table}
\centering
\caption{The 20dBZ CSI skill Score}\label{tab:csi}
\renewcommand\arraystretch{1.5}
\setlength\tabcolsep{0pt}
\begin{tabular}{c >{\bfseries}r @{\hspace{0.7em}}c @{\hspace{0.4em}}c }
  \multirow{12.5}{*}{\rotatebox{90}{\parbox{2cm}{\bfseries\centering Prediction}}} &
    & \multicolumn{2}{c}{\bfseries Ground Truth}  \\
  & & \bfseries $\geq20$dBZ & \bfseries $<20$dBZ  \\
  & $\geq20$dBZ & \MyBox{True}{Positive} & \MyBox{False}{Positive}  \\[2.4em]
  & $<20$dBZ & \MyBox{False}{Negative} & \MyBox{True}{Negative}

\end{tabular}
\end{table}
As an important criterion in precipitation nowcasting field, critical success index (CSI) skill score is defined as
\begin{equation}\label{csi}
    \mathrm{CSI} = \frac {\mathrm{hits}}{\mathrm{misses}+\mathrm{hits}+\mathrm{falsealarms}},
\end{equation}
where $\mathrm{hits}$ denotes the number of True Positives, in other words we predict a rainning area correctly. $\mathrm{misses}$ denote the number of False Negatives, which means we fail to forecast a raining area. $\mathrm{falsealarms}$ denote the number of False Positive as shown in Table \ref{tab:csi}.
\begin{equation}
    1-\mathrm{CSI} = \frac {\mathrm{misses}+\mathrm{falsealarms}}{\mathrm{misses}+\mathrm{hits}+\mathrm{falsealarms}},
\end{equation}
Intuitively, it might increase the CSI performance of forecasting by minimizing 1-CSI directly. However, CSI is not differentiable.

Alternatively, we approximate 1-CSI by
$\mathcal{L}_{\mathrm{SSL}}^{i}$ at classification point $c_i$:
\begin{equation}\label{eq:ms}
    \mathcal{L}_{\mathrm{SSL}}^{i}=\Big\|\sigma\Big((\mat{I} - c_i)*s\Big)-\sigma\Big((\widehat{\mat{I}}-c_i)*s\Big)\Big\|_2^2,
\end{equation}
where $s$ is the scale factor, a hyper-parameter to control the slope of the sigmoid function, and the subscript SSL is short for single sigmoid loss. $\|\cdot\|_2$ denotes the Frobenius norm.

$\mathcal{L}_{\mathrm{SSL}}$ is sensitive to falsealarms and misses,
less sensitive to others. The multi-sigmoid loss is composed of a
set of sigmoid losses
$\{\mathcal{L}_{\mathrm{SSL}}^{i}\}_{i=1,2,\cdots,n}$, in which
$\mathcal{L}_{\mathrm{SSL}}^{i}$ is to evaluate if $\mat{I}$ gives
out the correct classification for the classification point $c_i$:
\begin{equation}
  \mathcal{L}_{\mathrm{MSL}}=\sum_{i=1}^{n}\mathcal{L}_{\mathrm{SSL}}^{i}
\end{equation}
where the subscript MSL is short for multi-sigmoid loss.  In this work, we follow the weather forecasters' recommendation, $\{$20dBZ, 30dBZ, 40dBZ$\}$ in radar echo scale, as our critical points $\{c_1,c_2,\cdots,c_n\}$ for $\mathcal{L}_{\mathrm{MSL}}$. Scale factor $s$ is 15 based upon experiments.

\begin{figure}[t]
% \end{minipage}
% \begin{minipage}[t]{1\linewidth}\centering
\includegraphics[width=1\linewidth]{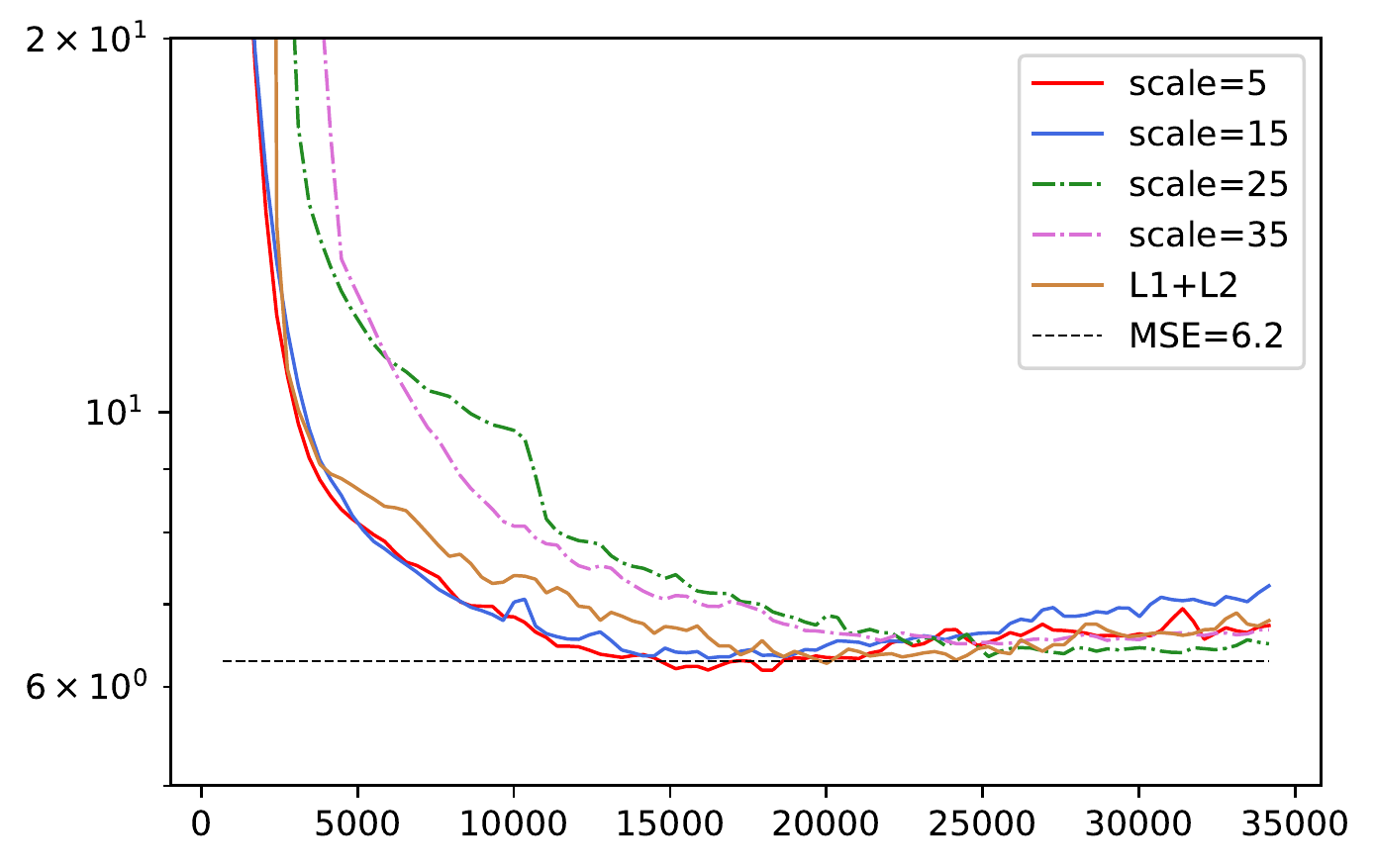}
% \end{minipage}
\vspace{-15px}
\caption{MSE performance  of different scale factors of $\mathcal{L}_{\mathrm{MSL}}$ }\label{fig:MSlossMSE}
\end{figure}
\section{Experiments} \label{experiment}
\subsection{Dataset and Metrics}
\paragraph{Dataset} Our radar echo dataset contains 170,000 weather radar intensity frames collected from October 2015 to July 2018 with the minimal interval of 6 minutes by the dual polarization Weather Surveillance Radar-1988 Doppler Radar (WSR-88D) located in Shanghai. Each frame is a 501 $\times$ 501 grid of image, covering almost 501 $\times$ 501 square kilometers. We normalize the echo intensity values $R$ ($0\leq R\leq 70$) into a gray scale $P$ by ${P = R / {70}}$. Our dataset is available  on Harvard Dataverse~\cite{DVN/2GKMQJ_2019}.

We applied a sliding window of stride 1 to generate 76,779  consecutive sequences of length 20. The training set (October 2015 to July 2017) and testing set (October 2017 to July 2018) contain 44,060 and 32,719 sequences, respectively. As a pre-processing step, we removed the rain-less frames from the training set to effectively train the network, while we kept the testing data unchanged. To fairly compare with other methods, we resized all the frames from 501 $\times$ 501 to 100 $\times$ 100.
\paragraph{Metrics.}
We report frame-wise mean square error (MSE) and CSI score for each experiment setting. Frame-wise MSE are calculated as follows,
\begin{equation}
    \mathrm{MSE} ={\frac {1}{L}}\sum _{t=1}^{L}\Big\|\mathrm{\mat{I}}_{t}-\widehat{\mat{I}}_{t}\Big\|_2^2,
\end{equation}
where $t$ is the index of radar echo frame and $L$ is the length of predicted frames. Another metric is the weather criterion, CSI, which has been defined in Eq.~\eqref{csi}. The critical point in CSI was set to be 20dBZ.

\begin{figure}[t]
\centering
% \begin{minipage}[t]{1\linewidth}
\includegraphics[width=1\linewidth]{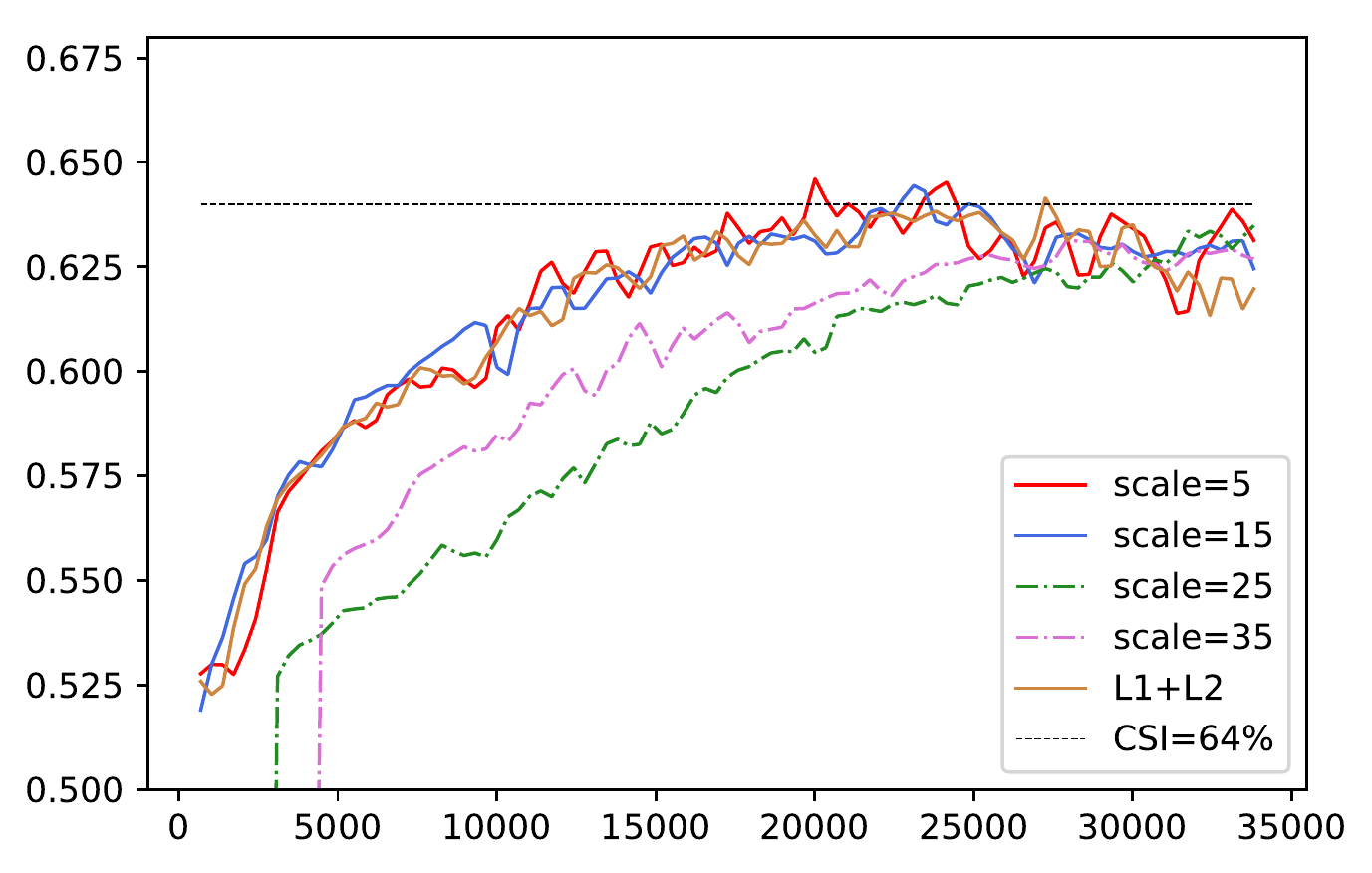}
\vspace{-15px}
\caption{CSI performance of different scale factors of $\mathcal{L}_{\mathrm{MSL}}$, X-axis stands for the number of training iteration.}\label{fig:MSlossCSI}
\centering
\end{figure}
\subsection{Implementation Details}
The lengths of both the input  and output sequence are 10. All our
experiments are implemented with PyTorch and conducted on 4 NVIDIA
GTX 1080Ti GPUs.  We trained our model with ADMM optimizer and the
learning rate is $0.002$. The size of mini-batch is $32$ and we stop
the training after 20,000 iterations. We use two stride-2 2-D
convolution layers to resize the input frames from
100$\times$100$\times$1 to 25$\times$25$\times$64. The kernel size
is (4,4). The number of channels of Group-Normalization are 16. Same
parameters are adopted for the transposed convolution layers. We use
a 2-layer encoding-forecasting structure withe the number of filters
set to 64, 64. The kernel size of StarBri LSTM is (3,3) .

\begin{figure*}[t]
\includegraphics[width=1.0\linewidth]{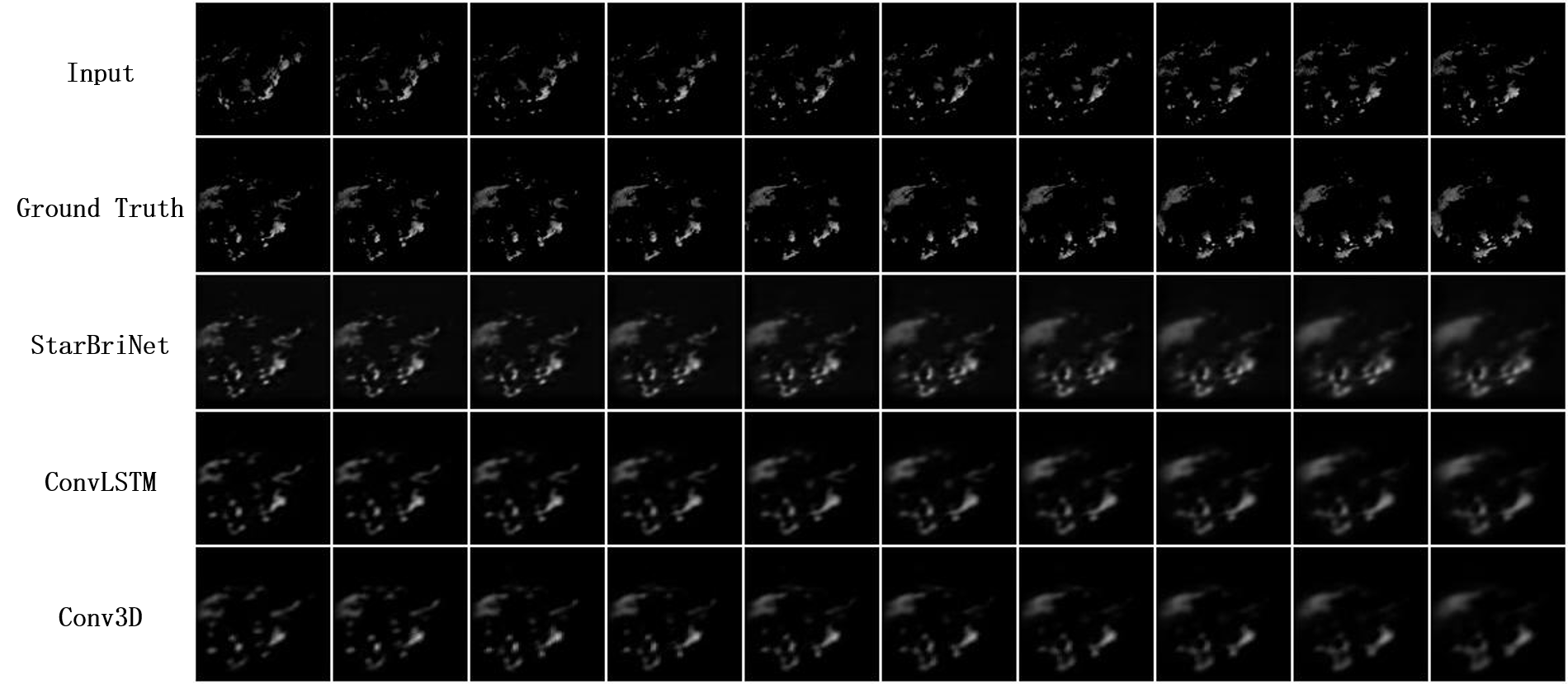}
\vspace{-15px}
\caption{Predictions on radar echo dataset with a interval of every 6 minutes.} \label{graysample}
\end{figure*}
\subsection{Experimental Results}
%TODO  the Frozen Prediciton which always output the latest frame,
The overall evaluation results are summarized in Table~\ref{tab1}.
We perform a detailed ablation study on multi-column encoder, star-shaped information bridge, and multi-sgimoid loss.
We also evaluate 5 other nowcasting algorithms, including 1 optical flow based methods (ROVER), 4 deep learning methods (ConvLSTM, PredRNN, TrajGRU, and 3D CNN). We implement a naive 8 layer 3D CNN, the kernel size is (3,3,3).
Both CSI (the higher the better) and MSE (the lower the better) increases  when we adpot multi sigmoid loss, because the multi sigmoid loss focus on misses and false-alarms, those are more correlated to CSI score. Overall, the StarBriNet model outperforms the other methods.
We visualize the generated radar maps in Fig.~\ref{graysample}. We can see that all methods generate the blurriness predictions. Because we treat precipitation nowcasting as a deterministic task. Unfortunately, this task are full of uncertainty. Our method treat this sample as a moderate one and generate more rainy area than others.

\begin{table}
\centering
\caption{Comparison results with other models on East-China radar echo dataset. MC is Multi-Column structure in section~\ref{3.1} and MSL is multi sigmoid loss in section~\ref{msl} }\label{tab1}
\begin{tabular}{lcc}
\toprule
Model & MSE & CSI [\%]\\
\midrule
ROVER &8.84 &58.8\\
3D CNN &6.33 &58.4 \\
ConvLSTM~\cite{convlstm} & 6.31 & 60.0 \\
TrajGRU~\cite{shi2017deep} & - & 63.0\\
PredRNN~\cite{wang2017predrnn} & 7.03 & 63.1 \\
\hline
ConvLSTM+MC & 6.51 & 62.1\\
StarBri LSTM+MC & \bfseries{6.12} & 63.8\\
StarBri LSTM+MC+MSL & 6.18 & \bfseries{64.4}  \\
\bottomrule
\end{tabular}
\end{table}

\subsection{Model Analysis}
\paragraph{Scale factors of $\mathcal{L}_{\mathrm{SSL}}^{i}$.}
We tested several scale factors in Fig.~\ref{fig:MSlossCSI} and
Fig.~\ref{fig:MSlossMSE}. Larger scale factor leads the gradient
sensitive to tiny differences at the starting phase of training
which influence the training robustness. To test our hypothesis, we
run an experiment of gradually increasing the scale factor from 1 to
40, with CSI score 64.27\%, MSE 6.17.

\paragraph{Resizing.}
Shi~\emph{et. al.}\cite{convlstm} first utilize a patch-resizing trick to transfer an input frame from 100$\times$100$\times$1 to 50$\times$50$\times$4 when patch-size is 2. This trick cut image into small patches and stack pixels inside each patch along channel-wise directly. The GPUs memory cosumtion and training time are greatly reduced during training. Wang~\emph{et. al.} ~\cite{predrnn,predrnnpp} also adopt this trick. However, combined with convolution layer in ConvLSTM layers, patch-resizing trick is equivalent to a big kernel size convolution. So we recommend to use alternate down-sampling and up-sampling strategies instead, e.g., the resize network in Fig.~\ref{fig:network}.

\paragraph{Data splitting strategy.}
Fig.~\ref{fig:Distribution} demonstrates the distribution of the
whole dataset, where each point in this figure represents for a
length-10 sample sequence of radar echo frames. Y-axis is the
changing rate of rain intensities in each sample sequence. Positive
value means the rain increased and negative means the rain is
letting up. X-axis is the average rainfall intensity of each sample.
Most samples lies in the green colored box, we category this as the
light rain cut, blue for the moderate cut and the purple area for
the heavy rain cut. This is a simple yet effective devision.

% \section{Dicussion}

\section{Conclusion}
We presented multi-column Star shape information Bridge Network (StrBriNet) for precipitation nowcasting task. a video prediction model that effectively passes feature and gradient through RNN layers. We proposed a new multi-sigmoid loss function which takes CSI score into consideration, and yield better prediction. We performed quantitative analysis on a Radar Echo dataset of east China. The results demonstrated that, compared with other method, our model outperform other state-of-the-art methods.

\end{document}